\PassOptionsToPackage{table}{xcolor}

\documentclass[10pt,twocolumn,letterpaper]{article}

\usepackage[pagenumbers]{cvpr} % To force page numbers, e.g. for an arXiv version

\usepackage{pifont}
\usepackage{multirow}
\usepackage{hhline}
\usepackage{booktabs}
\usepackage{listings}

\newcommand{\best}[1]{\textbf{\textcolor{red}{#1}}}
\newcommand{\secondbest}[1]{\underline{\textcolor{blue}{#1}}}
\newcommand\shline{\specialrule{0.8pt}{0pt}{0pt}}

\definecolor{cvprblue}{rgb}{0.21,0.49,0.74}
\usepackage[pagebackref,breaklinks,colorlinks,allcolors=cvprblue]{hyperref}
\makeatletter
\def\blfootnote{\xdef\@thefnmark{}\@footnotetext}
\makeatother

\title{OmniDrag: Enabling Motion Control for Omnidirectional \\Image-to-Video Generation}

\author{
\href{https://scholar.google.com/citations?user=SIkQdEsAAAAJ}{Weiqi Li}\textsuperscript{*1,2}, Shijie Zhao\textsuperscript{$\dagger$2}, Chong Mou\textsuperscript{1}, Xuhan Sheng\textsuperscript{1}, Zhenyu Zhang\textsuperscript{1}, \\ Qian Wang\textsuperscript{1}, Junlin Li\textsuperscript{2}, 
Li Zhang\textsuperscript{2}, 
\href{https://jianzhang.tech/}{Jian Zhang}\textsuperscript{$\dagger$1}\\
\textsuperscript{1} School of Electronic and Computer Engineering, Peking University,
\textsuperscript{2} ByteDance Inc \\
\url{https://lwq20020127.github.io/OmniDrag}
}

\begin{document}

\twocolumn[{
\renewcommand\twocolumn[1][]{#1}
\maketitle
\centering
\vspace{-0.8cm}
\includegraphics[width=\textwidth]{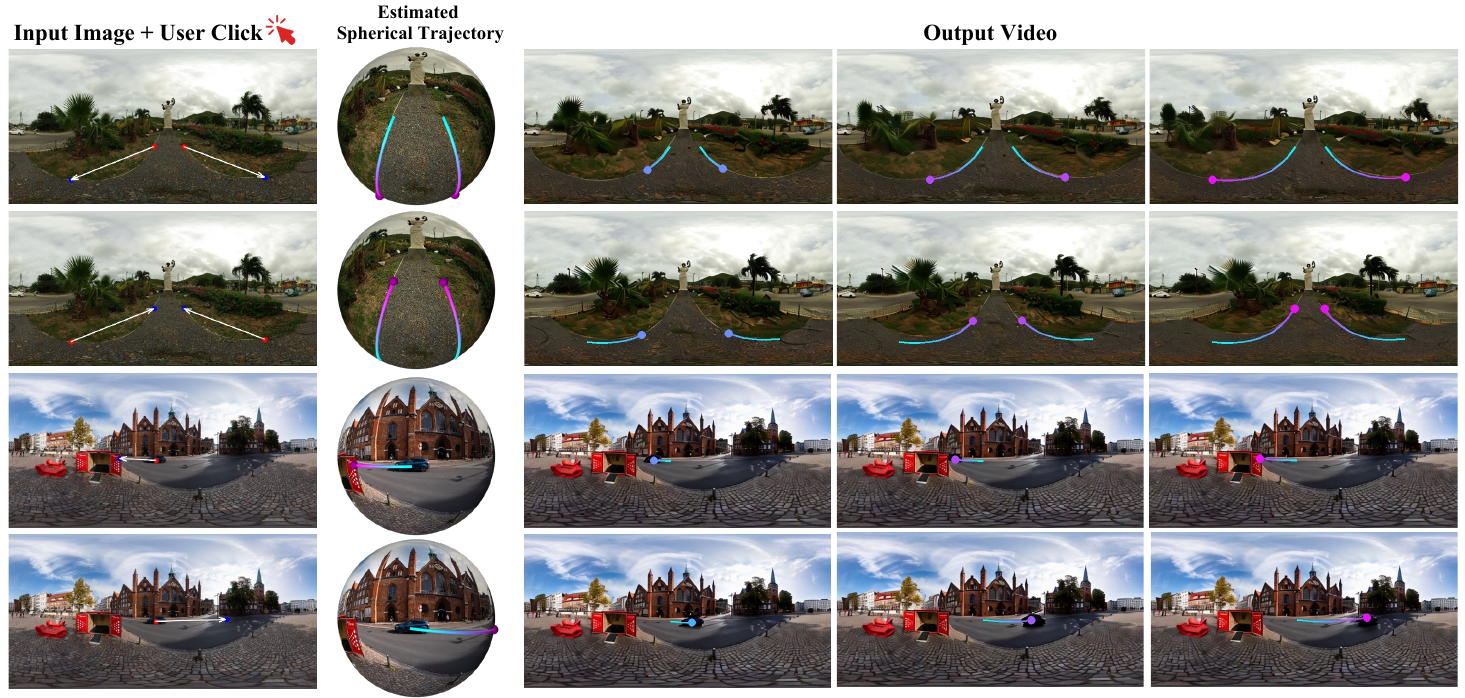}
\vspace{-0.7cm}
\captionsetup{type=figure}
\caption{\textbf{Omnidirectional videos generated by proposed OmniDrag.} It enables drag-style synthesis from a reference omnidirectional image and user-specified points, providing both scene-level (top) and object-level (bottom) accurate, high-quality controllable generation.}}\vspace{0.4cm}]

\begin{abstract}
As virtual reality gains popularity, the demand for controllable creation of immersive and dynamic omnidirectional videos (ODVs) is increasing. While previous text-to-ODV generation methods achieve impressive results, they struggle with content inaccuracies and inconsistencies due to reliance solely on textual inputs. Although recent motion control techniques provide fine-grained control for video generation, directly applying these methods to ODVs often results in spatial distortion and unsatisfactory performance, especially with complex spherical motions. To tackle these challenges, we propose \textbf{OmniDrag}, the first approach enabling both scene- and object-level motion control for accurate, high-quality omnidirectional image-to-video generation. \blfootnote{$^*$ This work was done during the internship at ByteDance.}\blfootnote{$^\dagger$ Corresponding author.}Building on pretrained video diffusion models, we introduce an omnidirectional control module, which is jointly fine-tuned with temporal attention layers to effectively handle complex spherical motion. In addition, we develop a novel spherical motion estimator that accurately extracts motion-control signals and allows users to perform drag-style ODV generation by simply drawing handle and target points. We also present a new dataset, named \textbf{Move360}, addressing the scarcity of ODV data with large scene and object motions. Experiments demonstrate the significant superiority of OmniDrag in achieving holistic scene-level and fine-grained object-level control for ODV generation. The project page is available at \url{https://lwq20020127.github.io/OmniDrag}. 
\end{abstract}

\vspace{-16pt}
\section{Introduction}
\label{sec:intro}
% \vspace{-4pt}
Omnidirectional video (ODV) \cite{xiao2012recognizing,zink2019scalable}, also known as 360° or panoramic video, has gained increasing attention due to its immersive and interactive capabilities, as well as its wide applications in virtual and augmented reality. It provides a full 360°$\times$180° field of view and is typically captured using an array of high-resolution fisheye cameras. Such a process is expensive in terms of both time and hardware resources in real-world scenarios \cite{ai2022deep}. Therefore, there is an urgent need for developing ODV generation methods.

In the field of 2D video generation, numerous diffusion-based models such as Gen-2 \cite{esser2023structure}, Stable Video Diffusion (SVD) \cite{blattmann2023stable}, and Sora \cite{videoworldsimulators2024} have achieved great success by leveraging powerful generative priors learned using large-scale training data and substantial computation resources. For ODV generation, 360DVD \cite{wang2024360dvd} introduces a plug-and-play 360-Adapter to enable text-to-ODV synthesis. However, this paradigm relies solely on text input, which often provides overly broad generation freedom and fails to precisely determine video frames, leading to inaccurate and inconsistent content control. While 360DVD offers optical flow-based control, obtaining ODV optical flow is challenging for users \cite{shi2023panoflow}, thus limiting its practical utility.

Recently, trajectory-based motion control has emerged as a more user-friendly and effective solution for controllable video generation. Drawing trajectories offers a simple yet flexible approach, compared to other control signals like optical flow or depth maps \cite{hao2018controllable}. Based on this approach, efforts such as DragNUWA \cite{yin2023dragnuwa}, MotionCtrl \cite{wang2024motionctrl}, and DragAnything \cite{wu2025draganything} encode sparse trajectories or camera motions into latent space to effectively guide object movements. Despite these advanced methods for 2D video synthesis, directly applying them to ODV generation presents three significant challenges: \textit{\textbf{Firstly,}} unlike controlling traditional 2D videos, which generally involve simple motions, the motion patterns in ODVs are often spherical. Previous approaches applied in this task can lead to spatial distortions in generated results due to their inability to model complicated spherical motions. \textit{\textbf{Secondly,}} since ODVs are generally stored in equirectangular projection (ERP) format, controlling them is more difficult than controlling 2D videos, as drawing reasonable and precise spherical motion trajectories on ERP images is challenging for human users. \textit{\textbf{Thirdly,}} existing ODV datasets contain samples with limited motion magnitudes, constraining the effectiveness of deep controllable ODV generation models when faced with users' requirements for larger motion ranges.

To address these problems, in this paper, we propose \textbf{OmniDrag}, the first method to enable motion control for omnidirectional image-to-video generation based on powerful pretrained video diffusion models. As demonstrated in Fig.~\textcolor{cvprblue}{1}, OmniDrag achieves high-quality, controllable ODV generation with simple user input, enabling both scene-level and object-level drag-style control using a unified model. In OmniDrag, we introduce an omnidirectional controller that takes trajectory as input to provide fine-grained motion controllability. To effectively learn complicated spherical motions in ODVs, we propose jointly fine-tuning the temporal attention components with our controller. For accurate and easy motion control, we develop a novel spherical motion estimator (SME). During training, SME tracks object motion using an equal-area iso-latitude spherical point initialization \cite{gorski2005healpix} and samples through a filter based on spherical distance to capture important movements uniformly and accurately. During inference, SME estimates motion trajectories via spherical interpolation, allowing users to provide only the handle and target points. Furthermore, we introduce a new high-quality ODV dataset named \textbf{Move360}, featuring significant scene-level and object-level motions. Move360 comprises more than 1,500 video clips across diverse scenes, captured by an Insta360 Titan mounted on a filming car. Experiments show that training on Move360 enhances OmniDrag's ability for scene-level movement.

In summary, our contributions are:
\begin{itemize}
\item We propose \textbf{OmniDrag}, a novel method enabling motion control for ODV generation. It learns spherical motion patterns by jointly fine-tuning an omnidirectional controller and temporal attention layers in the UNet denoiser.
\item  We develop a novel spherical motion estimator (SME) that accurately captures control signals during training and allows users to simply draw handle and target points during inference, providing user-friendly controllability.
\item  We introduce \textbf{Move360}, a new high-quality ODV dataset, featuring large camera and object movements in samples captured by an Insta360 Titan mounted on a filming car, which enhances OmniDrag's scene-level controllability.
\item  Extensive experiments demonstrate OmniDrag's effectiveness and superior performance in generating smooth and visually appealing ODVs under interactive motion control, including both scene- and object-level control.
\end{itemize}

\vspace{-6pt}
\section{Related Work}
\label{sec:related}

\begin{figure*}[!ht]
\includegraphics[width=\textwidth]{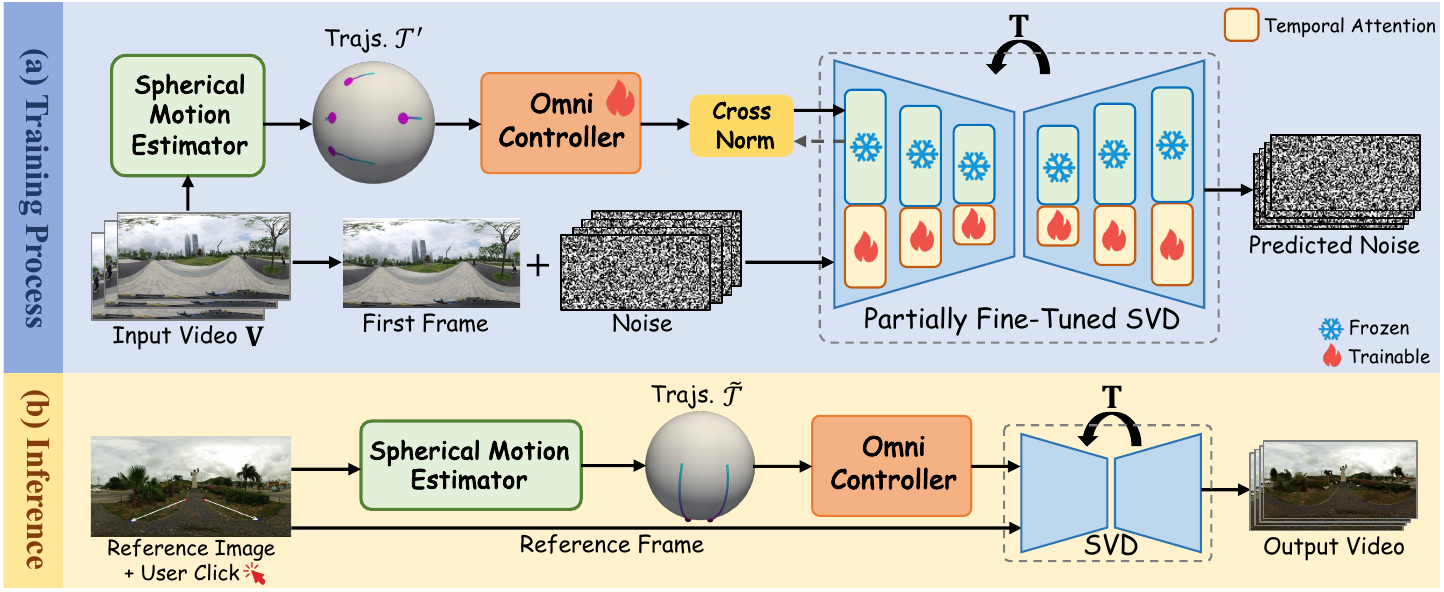}
\vspace{-16pt}
\caption{\textbf{Overall pipeline of proposed OmniDrag.} \textbf{\textcolor{blue}{(a)}} During training, spherical motion is extracted by the proposed spherical motion estimator. The Omni Controller and temporal attention layers in the UNet denoiser are jointly fine-tuned. \textbf{\textcolor{blue}{(b)}} During inference, OmniDrag allows users to simply select handle and target points on the reference image and generates ODVs with the corresponding motion.}
\vspace{-12pt}
\label{fig:pipeline}
\end{figure*}

\subsection{Controllable Image and Video Generation}

Recent developments in diffusion models \cite{ho2020denoising, dhariwal2021diffusion} have significantly enhanced image and video generation capabilities. Leading image generation frameworks, such as Stable Diffusion \cite{rombach2022high}, Imagen \cite{saharia2022photorealistic}, and DALL-E2 \cite{ramesh2022hierarchical}, utilize textual inputs to guide the generation process. Approaches like ControlNet \cite{zhang2023adding} and T2I-Adapter \cite{mou2024t2i} incorporate additional control modules into these pre-trained diffusion models to achieve finer controllability. In video generation, early methods similarly depend on text-based conditions, as demonstrated by Video LDM \cite{blattmann2023align}, Imagen Video \cite{ho2022imagen}, and AnimateDiff \cite{guo2023animatediff}. However, text prompts often fall short in handling complex scenarios, prompting recent research \cite{blattmann2023stable, esser2023structure, xing2023dynamicrafter, zhang2023i2vgen} to adopt image-based conditions for more precise and effective control. For example, Video ControlNet \cite{chen2023control,zhang2023controlvideo} extends the ControlNet architecture to video generation by conditioning on sequences of control signals such as depth and edge maps. ControlNeXt \cite{peng2024controlnext} further improves ControlNet for lightweight image and video control guidance. Effective motion control is essential for producing coherent and dynamic videos. Current strategies employ trajectory-based methods like DragNUWA \cite{yin2023dragnuwa}, MotionCtrl \cite{wang2024motionctrl}, DragAnything \cite{wu2025draganything}, and Tora \cite{zhang2024tora}, as well as box-based techniques \cite{ma2023trailblazer, jain2024peekaboo, wang2024boximator, qiu2024freetraj}. Unlike these 2D video generation methods that achieve desired motion dynamics by training additional motion controllers on frozen pre-trained video diffusion models, OmniDrag focuses on learning complex spherical motion patterns by jointly fine-tuning temporal attention layers in the base SVD model.

\subsection{Omnidirectional Image and Video Generation}
Generative adversarial network-based methods for producing omnidirectional images (ODIs) have been extensively explored \cite{oh2022bips, teterwak2019boundless, wu2022cross, lin2019coco, lin2021infinitygan, cheng2022inout, wang2022stylelight, chen2022text2light, dastjerdi2022guided, akimoto2022diverse, ai2024dream360,li2024scenedreamer360}. Recently, diffusion models have significantly advanced ODI generation \cite{zhang2023diffcollage,li2023panogen,wang2023360,wang2023customizing, wu2023panodiffusion, zhang2024taming, yang2024layerpano3d, li20244k4dgen,li2024d3c2,liu2021hybrid,yang2024fourier123,chen2024invertible,zhang2024editguard,zhang2024v2a}. Specifically, PanoDiffusion \cite{wu2023panodiffusion} employs a dual-modal diffusion architecture incorporating RGB-D data to capture the spatial patterns of ODIs. PanFusion \cite{zhang2024taming} introduces a dual-branch diffusion model that integrates global panorama and local perspective latent domains. LayerPano3D \cite{yang2024layerpano3d} decomposes a reference ODI into multiple layers at varying depth levels to facilitate explorable panoramic scenes. In the realm of omnidirectional video (ODV) generation, 360DVD \cite{wang2024360dvd} utilizes motion modeling modules \cite{guo2023animatediff} and 360Adapter to enable text-to-ODV generation with optical flow control. DiffPano~\cite{ye2024diffpano} introduces a spherical epipolar-aware multi-view diffusion model. However, relying solely on text inputs often leads to inaccuracies and inconsistencies in the generated frames, and acquiring ODV optical flow poses challenges for users, limiting broader applications. In contrast, our OmniDrag enables control through images and trajectories, providing accurate controllability with a user-friendly interface.

% \vspace{-6pt}
\section{Methodology}
In this section, we begin with a concise review of the employed base model Stable Video Diffusion (Sec.~\ref{sec:3.1}). Following this, we provide an overview of our OmniDrag (Sec.~\ref{sec:3.2}), illustrated in Fig.~\ref{fig:pipeline}. We then elaborate on the Omni Controller and partial fine-tuning technique (Sec.~\ref{sec:3.3}) and proposed spherical motion estimator (Sec.~\ref{sec:3.4}). The proposed Move360 dataset is detailed in Sec.~\ref{sec:3.5}.

\subsection{Preliminaries}
\label{sec:3.1}
\textbf{Stable Video Diffusion} (SVD) \cite{blattmann2023stable} is a high-quality and widely used image-to-video generation model. We adopt SVD as the base model for our proposed OmniDrag, to leverage its high-quality video generation capabilities. Specifically, given a reference image $\mathbf{c}_I$, SVD generates a sequence of video frames of length $L$, starting with given $\mathbf{c}_I$, denoted as $\mathbf{x} = \{\mathbf{x}^0, \mathbf{x}^1, \dots, \mathbf{x}^{L-1}\}$. Following the latent denoising diffusion process in \cite{rombach2022high}, a 3D UNet $\mathbf{\Phi}_\theta$ is used to denoise the sequence iteratively at timestep $t$:
\begin{equation}
\hat{\mathbf{z}}_0=\mathbf{\Phi}_\theta(\mathbf{z}_t, t, \mathbf{c}_I),
\end{equation}
where $\mathbf{z}_t$ is the latent representation of $\mathbf{x}_t$ obtained via an autoencoder \cite{van2017neural,jia2023taming} as $\mathbf{z}_t=\mathcal{E}(\mathbf{x}_t)$, and  $\hat{\mathbf{z}}_0$ is the model's prediction of $\mathbf{z}_0=\mathcal{E}(\mathbf{x})$. To inject the reference image $\mathbf{c}_I$ into the main denoising branch, there are two paths: (1) $\mathbf{c}_I$ is embedded into tokens by the CLIP \cite{radford2021learning} image encoder and injected into the diffusion model through a cross-attention \cite{rombach2022high} mechanism. (2) $\mathbf{c}_I$ is encoded into latent representation by the VAE encoder \cite{van2017neural,jia2023taming} of the latent diffusion model and concatenated with the latent representations of each frame along the channel dimension. SVD parameterizes the learnable denoiser $\mathbf{\Phi}_\theta$ following the EDM-preconditioning \cite{karras2022elucidating} framework, as:
\begin{equation}
\begin{aligned}
\mathbf{\Phi}_\theta(\mathbf{z}_t&,t, \mathbf{c}_I;\sigma)=c_{{skip}}(\sigma)\mathbf{z}_t \\
+&c_{{out}}(\sigma)F_\theta(c_{{in}}(\sigma)\mathbf{z}_t, t, \mathbf{c}_I;c_{{noise}}(\sigma)),
\end{aligned}
\end{equation}
where $\sigma$ is the noise level, $F_\theta$ is the denoising network, and $c_{skip}$, $c_{out}$, $c_{in}$ and $c_{noise}$ are hyper-parameters conditioned on $\sigma$. Finally, $\mathbf{\Phi}_\theta$ is trained via denoising score matching:
\begin{equation}
    \mathbb{E}_{\mathbf{z}_0, t, \mathbf{n}\sim\mathcal{N}(0, \sigma^2)}\left[\lambda_\sigma||\mathbf{\Phi}_\theta(\mathbf{z_0}+\mathbf{n},t,\mathbf{c}_I)-\mathbf{z}_0||^2_2\right].
\end{equation}

\subsection{Overview of OmniDrag}
\label{sec:3.2}
An overview of OmniDrag is illustrated in Fig.~\ref{fig:pipeline}. Built upon the pretrained SVD model, OmniDrag operates as follows. During training, the proposed spherical motion estimator (SME) first extracts trajectories from the input video. These trajectories are then fed into the Omni Controller, which is jointly fine-tuned with the temporal attention layers of the U-Net denoiser. During inference, users can simply select handle and target points on a reference image. OmniDrag then generates ODVs exhibiting the corresponding motion, enabling intuitive and precise motion control.

\subsection{Omni Controller and Partial Fine-Tuning}
\label{sec:3.3}

\textbf{Motivation.} Temporal attention layers play important roles in motion pattern learning for diffusion models \cite{ku2024anyv2v, bai2024uniedit, hu2024motionmaster}. Existing 2D motion-guided methods \cite{yin2023dragnuwa, wu2025draganything, wang2024motionctrl} freeze the main branch of UNet model and utilize a trainable copy of the UNet encoder to inject the motion control. However, different from 2D videos, which generally involve simple motions, the motion patterns in ODVs are often spherical, introducing a significant gap. Consequently, merely training a control module with frozen main UNet branch leads to output videos with spatial distortions (Fig.~\ref{fig:ablation_temporal} in Sec.~\ref{sec:exp}). Therefore, we leverage a lightweight Omni Controller and jointly fine-tune the temporal attention layers in the denoising UNet to effectively learn the spherical motion pattern.

\textbf{Method.} Our OmniDrag employs a lightweight Omni controller instead of using a fully trainable copy of the main UNet encoder. Specifically, inspired by the recent ControlNeXt \cite{peng2024controlnext}, we use a lightweight convolutional module only consisting of multiple ResNet blocks \cite{he2016deep} to extract control signals. These controls are then integrated into the main denoising branch at a single selected middle block via an addition operation. Mathematically, 
\begin{equation}
\begin{aligned}
\label{eq:controller}
\mathbf{y}_m = \mathcal{F}_m\left(\mathbf{z}, \mathcal{F}_{c}(\mathbf{c}; \Theta_c);\Theta_m\right),
\end{aligned}
\end{equation}
 where $\mathbf{y}_m$ represents the updated diffusion feature, $\mathcal{F}_m$ and $\mathcal{F}_c$ denote the main denoising U-Net and the Omni controller with parameters $\Theta_m$ and $\Theta_c$, respectively. We propose to jointly fine-tune the temporal attention layers in the main UNet branch, whose parameters are denoted as $\Theta_t \subseteq \Theta_m$, as depicted in Fig.~\ref{fig:pipeline}. This joint fine-tuning process is crucial for learning spherical motion patterns, resulting in the parameter set of OmniDrag $\Theta=\{\Theta_c, \Theta_t\}$. Additionally, we adopt the cross-normalization \cite{peng2024controlnext} technique to efficiently inject motion control signals into the main UNet branch during the fine-tuning process, aligning the distributions of the denoising and control features. Denoting the latent condition signals as $\mathbf{z}_c=\mathcal{F}_c(\mathbf{c};\Theta_c)$, the final normalized control $\hat{\mathbf{z}}_c$ is calculated as:
\begin{equation}
\begin{aligned}
\label{eq:cross_norm}
\hat{\mathbf{z}}_c=\frac{\mathbf{z}_c-\pmb{\mu}_m}{\sqrt{\pmb{\sigma}^2_m + \epsilon}} * \gamma,
\end{aligned}
\end{equation}
where $\pmb{\mu}_m$ and $\pmb{\sigma}_m$ are the mean and variance of the latent $\mathbf{z}$ from the main branch, respectively. $\gamma$ is a hyper-parameter to scale the normalized value, and $\epsilon$ is a small constant for numerical stability. $\hat{\mathbf{z}}_c$ is finally integrated into the main denoising branch through addition.
More details of the Omni Controller are provided in the supplementary materials.

\begin{figure}[!t]
    \centering
    \includegraphics[width=1.0\linewidth]{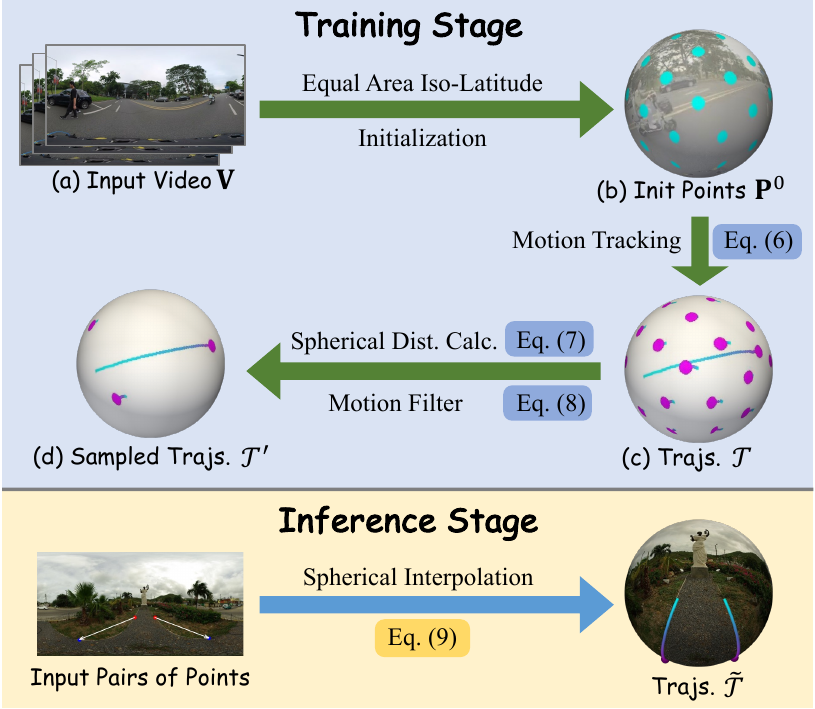}
    \vspace{-14pt}
    \caption{\textbf{Illustration of our spherical motion estimator (SME).} In the training stage, given the input video $\mathbf{V}$, $\mathbf{P}^0$ is firstly initialized through equal area iso-latitude pixelation. Then trajectories $\mathcal{T}$ are tracked, and finally filtered as $\mathcal{T}'$ according to spherical distance via Eqs.~(\ref{eq:track}-\ref{eq:filter}). During inference, given point pairs by users, the trajectories are estimated through spherical interpolation.}
    \label{fig:SME}
    \vspace{-12pt}
\end{figure}

\begin{figure*}[!t]
    \centering
    \includegraphics[width=1.\linewidth]{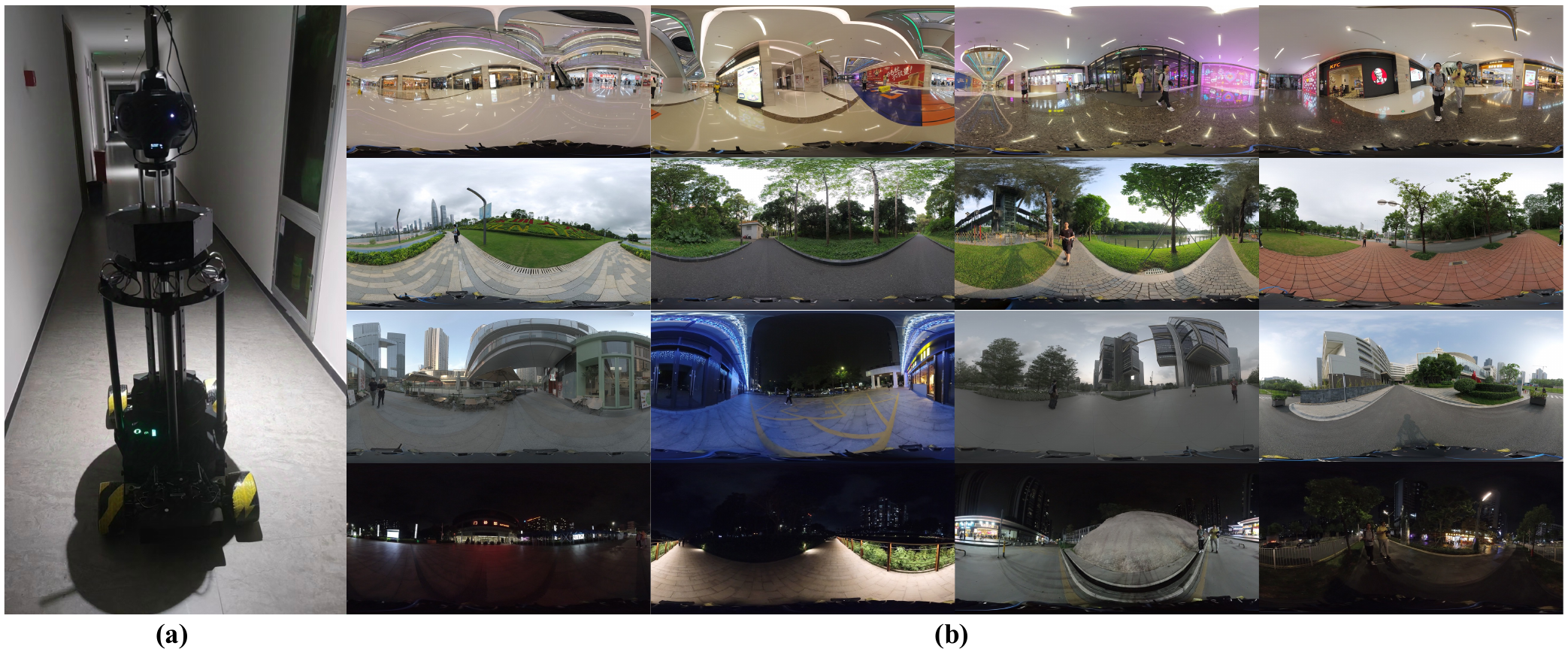}
    \vspace{-18pt}
    \caption{\textbf{Our Move360 dataset.} \textbf{\textcolor{blue}{(a)}} We mount Insta360 Titan on a filming car, enabling its movement along four degrees of freedom. \textbf{\textcolor{blue}{(b)}} Sample frames from the Move360 dataset showcasing a wide range of scenes, including indoor spaces, green landscapes, urban environments, and nighttime settings.  This diversity in motion and environments offers a rich dataset for the community.}
    \label{fig:move360}
    \vspace{-12pt}
\end{figure*}

\subsection{Spherical Motion Estimator}
\label{sec:3.4}
\textbf{Motivation.} Precise motion control signals are essential for both training and inference phases. Existing 2D motion control methods typically initialize tracking points on images using uniform grids \cite{yin2023dragnuwa, wu2025draganything, mou2024revideo} and perform probability sampling based on motion distance. However, due to the spatial distortion inherent in equirectangular projection (ERP)~\cite{li2024resvr,sun2023opdn,li2025omnissr, cao2023ntire,cheng2023hybrid}, pixel density decreases near the poles, resulting in inaccurate and oversampled motion tracking in these areas. Moreover, directly calculating distances on the ERP does not reflect true spherical motion magnitudes, often causing primary motion patterns to be overlooked. Additionally, during inference, current methods require users to manually draw complete trajectories, which is challenging for users to draw reasonable spherical paths on the ERP image. To overcome these limitations, we propose the spherical motion estimator (SME). As illustrated in Fig.~\ref{fig:SME}, SME captures more accurate spherical motion trajectories during training and offers user-friendly control capabilities during inference.

\begin{figure*}[!ht]
    \centering
    \includegraphics[width=\linewidth]{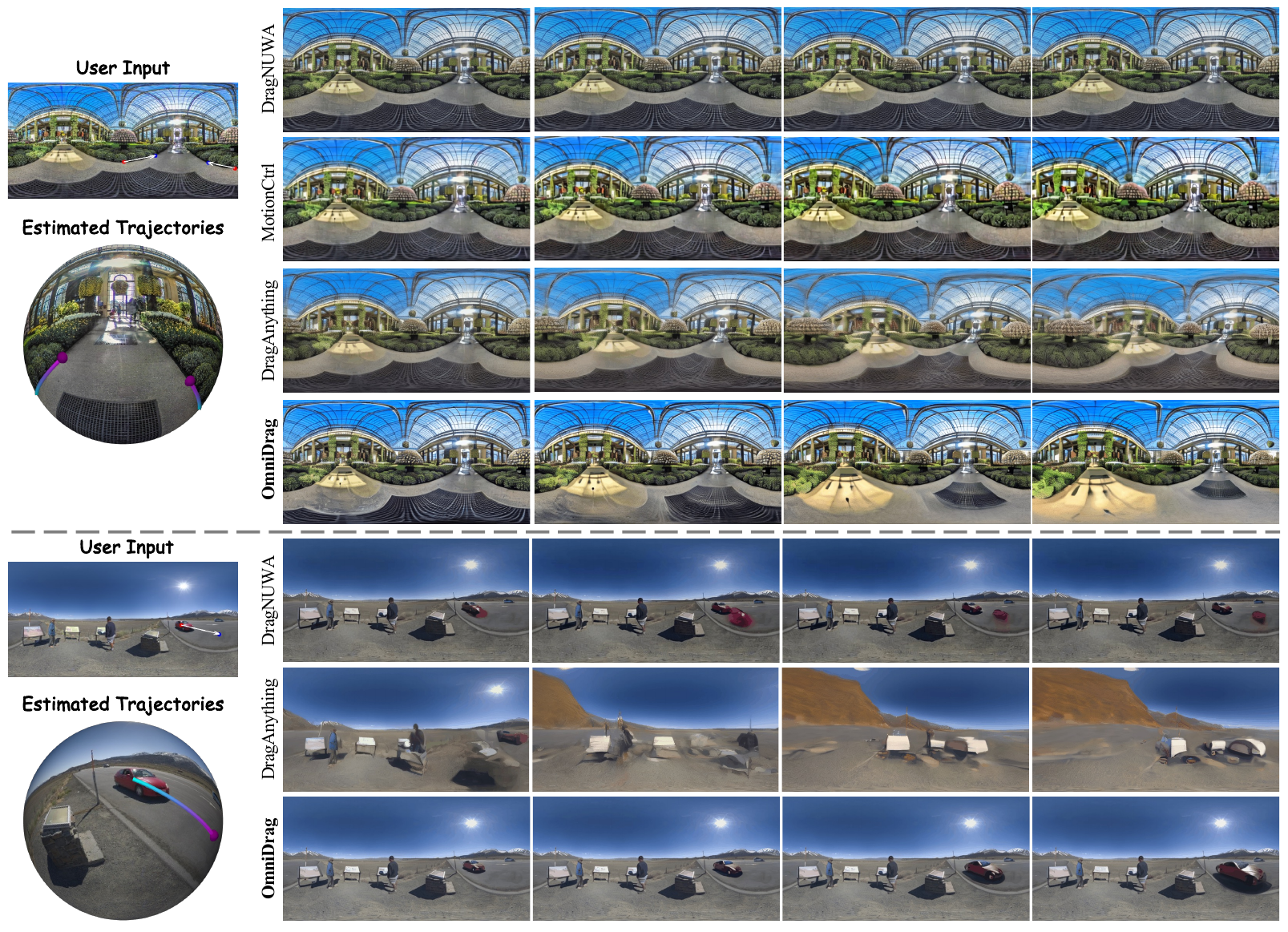}
    \vspace{-20pt}
    \caption{\textbf{Visual comparisons} between DragNUWA \cite{yin2023dragnuwa}, MotionCtrl \cite{wang2024motionctrl}, DragAnything \cite{wu2025draganything}, and our OmniDrag. Our SME estimates reasonable trajectories on the sphere, and OmniDrag achieves precise and stable control under both \textcolor{blue}{scene-level} (the \textcolor{blue}{top} case: go forward on the road) and \textcolor{blue}{object-level} (the \textcolor{blue}{bottom} case: make the car move along the road) motion conditions, outperforming other methods.}
    \label{fig:comparison}
\end{figure*}

\textbf{Method.} During training, we extract motion trajectories $\mathcal{T}$ from the input video to generate motion conditions. Let $N_{init}$ denote the number of initialized tracking points and $L$ the length of the video. Each trajectory $\mathbf{T}_j\in \mathcal{T}$ is defined as a sequence of spatial positions $\mathbf{T}_j=\{(x^i_j, y^i_j)|i\in \{0, 1, \dots, L-1\}\}$, where $(x_j^i, y_j^i)$ represents the position of the $j$-th trajectory at frame $i$.  To uniformly sample trajectories on the sphere, we propose to initialize the tracking points using the hierarchical equal area iso-latitude pixelation (HEALPix) grid \cite{gorski2005healpix}, which provides a uniform distribution of grid points on the sphere, assigning the same area to each pixel, as shown in Fig.~\ref{fig:SME}. Specifically, given a resolution parameter $N_{side}$, the HEALPix coordinate mapping function outputs a set of initialized points at frame $0$, as $\mathbf{P}^0=\{(x^0_j, y^0_j) |j \in \{0, 1, \dots, N_{init}-1 \}\}$, where the total number of points $N_{init}$ is $N_{init} = 12 \times {N_{side}}^2$. An object tracking function $\mathcal{F}_t$~\cite{karaev2023cotracker} is then applied to track the motion of these initialized points $\mathbf{P}^0$ across the input video $\mathbf{V} \in \mathbb{R}^{L\times C \times H \times W}$, generating the corresponding motion trajectories as:
\begin{equation}
\label{eq:track}
\begin{aligned}
    \mathcal{T} = \mathcal{F}_t\left(\mathbf{P}^0, \mathbf{V}\right),
\end{aligned}
\end{equation}
where $\mathcal{T}  \in \mathbb{R}^{N_{init}\times L \times 2}=\{\mathbf{T}_j|j \in \{0, 1, \dots, N_{init}-1\}\}$. Trajectories exhibiting larger motions are particularly beneficial for learning motion controllability. Therefore, we select trajectories with greater motion magnitudes. Instead of measuring motion magnitude in the ERP format, we propose to identify the primary motion of ODVs based on spherical distance, which is calculated as:
\begin{equation}
\begin{aligned}
\label{eq:spherical_dist}
    D(\mathbf{T}_j) =& \arccos \left( \sin(\theta^0_j) \sin(\theta^{L-1}_j) \right. \\
    &+\left.\cos(\theta^0_j) \cos(\theta^{L-1}_j) \cos(\phi^0_j - \phi^{L-1}_j) \right),
\end{aligned}
\end{equation}
where the angles $\phi$ and $\theta$ are obtained by converting the ERP points $(x, y)$ to spherical coordinates using $\phi = {2\pi x}/{W} - \pi$ and $\theta = {\pi y}/{H} - {\pi}/{2}$. We then filter $\mathcal{T}$ as:
\begin{equation}
\label{eq:filter}
\begin{aligned}
    \mathcal{T}'=\{\mathbf{T}\in\mathcal{T}|D(\mathbf{T}_j)>d_{th}\},
\end{aligned}
\end{equation}
where $d_{th}$ is a threshold. Following \cite{mou2024revideo}, we use the normalized distances as sampling probabilities to randomly select $N_{samp}$ trajectories from $\mathcal{T}'$. The final condition map $\mathbf{c}$ is then obtained by applying a Gaussian filter to smooth the sampled trajectories. Consequently, $\mathbf{c}$ serves as the conditional input in Eq.~(\ref{eq:controller}) to guide the generation process.

During inference, our objectives are: (1) to provide user-friendly interaction, and (2) to align the control signals with those used during training. Existing methods require users to draw motion trajectories on the reference image, which is feasible on 2D planar images. However, due to the spatial distortions of the ERP format, it is challenging for users to draw accurate spherical paths on the reference ERP image. To address this challenge, we introduce an innovative approach where users only need to specify the handle and target points, and the entire trajectory is then automatically estimated through spherical interpolation. Mathematically, denoting a pair of handle and target points as $(x^0, y^0)$ and $(x^{L-1}, y^{L-1})$, these points are firstly transformed to spherical coordinates $(\theta^0, \phi^0)$ and $(\theta^{L-1}, \phi^{L-1})$. Then, the intermediate points $(\theta^i, \phi^i)$ are calculated as:
\begin{equation}
\label{eq:slerp}
\left\{
\begin{aligned}
    \theta^i &= \scalebox{1.0}{$
    \arcsin \left( \frac{\sin\left((1 - t_i) \omega\right) \sin\theta^0 + \sin\left(t_i \omega\right) \sin\theta^{L-1}}{\sin\omega} \right)
    $} \\
    \phi^i &= \phi^0 + t_i \left( \phi^{L-1} - \phi^0 \right),
\end{aligned}
\right.
\end{equation}
where $\omega$ is the spherical distance between these two points calculated as Eq.~(\ref{eq:spherical_dist}), and $t_i = i / L$ is the interpolation factor. Finally, these points are transformed back to ERP coordinates, and combined with the handle and target points to obtain $\widetilde{\mathcal{T}} \in \mathbb{R}^{N_{p}\times L \times 2}$, where $N_{p}$ is the number of point pairs provided by the user. This process aligns inference-time control signals with those used during training.

\begin{table*}[!ht]
\caption{\textbf{Quantitative comparisons} between our OmniDrag and other methods. We employ automatic metrics (FVD \cite{unterthiner2018towards}, FID \cite{seitzer2020pytorch} and ObjMC \cite{wang2024motionctrl}) on both ERP format and final horizontal eight viewports. We also conduct a human evaluation to assess the performance. Throughout this paper, the best and second-best results are highlighted in \best{bold red} and \secondbest{underlined blue}, respectively.}
\vspace{-6pt}
\label{tab:comparison}
\centering
\small
\resizebox{0.8\linewidth}{!}{
\begin{tabular}{l|ccc|cc|cc}
\shline
\rowcolor[HTML]{EFEFEF} &
  \multicolumn{3}{c|}{\small\cellcolor[HTML]{EFEFEF}ERP Image} &
  \multicolumn{2}{c|}{\small\cellcolor[HTML]{EFEFEF}Horizontal 8 viewports} & \multicolumn{2}{c}{\small\cellcolor[HTML]{EFEFEF} Human Evaluation} \\ \hhline{>{\arrayrulecolor[HTML]{EFEFEF}}->{\arrayrulecolor{black}}|-------} 
\rowcolor[HTML]{EFEFEF} 
\multicolumn{1}{c|}{\multirow{-2}{*}{\cellcolor[HTML]{EFEFEF}Method}} & 
 \small FID$\downarrow$ &
 \small FVD$\downarrow$ &
 \small ObjMC$\downarrow$ &
 \small FID$\downarrow$ &
 \small FVD$\downarrow$ &
 \small Overall $\uparrow$ &
  \small Motion Matching $\uparrow$
  \\ \hline \hline

{DragNUWA \cite{yin2023dragnuwa}}  & \best{164.84} & \secondbest{1015.32} & 0.418 & \secondbest{96.31} & \secondbest{379.16} & \secondbest{15.0\%} & 9.5 \% \\ 

{DragAnything \cite{wu2025draganything}}  & 182.63 & 1113.95 & \secondbest{0.085} & 109.76 & 401.43 & 9.3\% & \secondbest{14.4\%}\\ 

\textbf{OmniDrag (Ours)} & \secondbest{171.41} & \best{933.73} & \best{0.044} 
& \best{95.62} & \best{322.22} &  \best{75.7\%} & \best{76.1\%} \\ \shline
\end{tabular}}
\vspace{-10pt}
\end{table*}

\subsection{Move360 Dataset}
\label{sec:3.5}
\begin{figure}[!t]
    \centering    \includegraphics[width=1.0\linewidth]{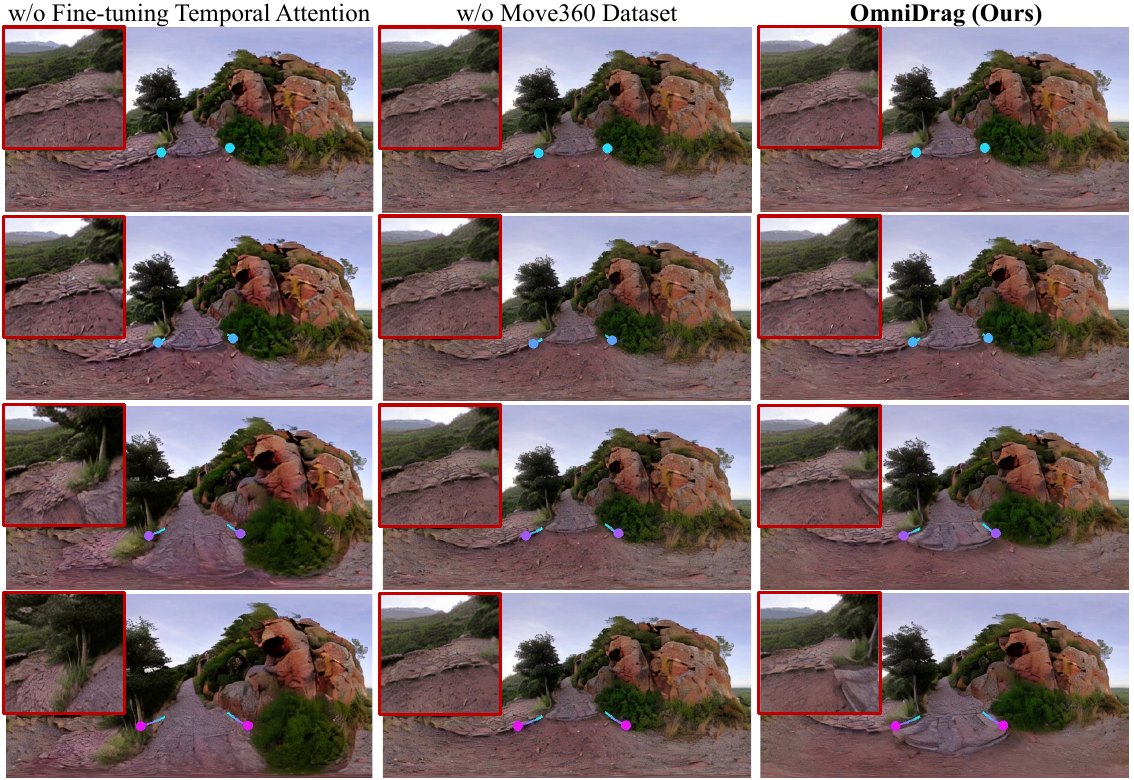}
    \vspace{-16pt}
    \caption{\textbf{Ablation study} on jointly fine-tuning temporal attention layers, and training with proposed Move360 dataset. For each ERP image, we show a corresponding viewport at specific perspective.}
    \label{fig:ablation_temporal}
    \vspace{-10pt}
\end{figure}

Training OmniDrag requires ODV datasets with high-quality motion. However, existing ODV datasets offer limited motion quality and magnitude due to their data acquisition methods. Specifically, videos in WEB360 \cite{wang2024360dvd} are primarily collected from the ``AirPano VR" YouTube channel. These videos are obtained through aerial photography and contain watermarks, resulting in limited motion patterns and quality. The 360+x dataset \cite{chen2024360+} includes multiple scenes of ODVs from a third-person perspective. However, most videos in 360+x are filmed with a stationary camera, which is not conducive to learning motion. As shown in Fig.~\ref{fig:ablation_temporal}, training with existing datasets results in OmniDrag lacking scene-level control capabilities. To address these issues and cover the absence of high-quality panoramic video datasets with large motions, we introduce a new ODV dataset named Move360. Specifically, we mount an Insta360 Titan camera on a filming car. The Insta360 Titan features eight 200°F3.2 fisheye cameras. The captured circular videos are subsequently de-warped and stitched using optical flow. Ultimately, we obtained ODVs at a resolution of $7680\times 3840$ (8K) with a frame rate of 30 FPS. Moreover, our filming car allows the camera to move forward and backward, left and right, up and down, and rotate 360 degrees horizontally. These four degrees of freedom provide flexibility in capturing immersive content from various angles and positions. The original video has a duration of approximately 20 hours with 6TB in size. We curated the data based on scene and content quality, resulting in 1,580 clips from over 60 scenes, each consisting of 100 frames. Move360 contains a wide range of scenes as shown in Fig.~\ref{fig:move360}, offering a rich dataset for training models requiring high-quality ODV content. More video samples from Move360 are provided in the supplementary materials.

\section{Experiments}
\label{sec:exp}
\subsection{Experimental Setup}

\begin{figure*}[!ht]
    \centering
    \includegraphics[width=1.0\linewidth]{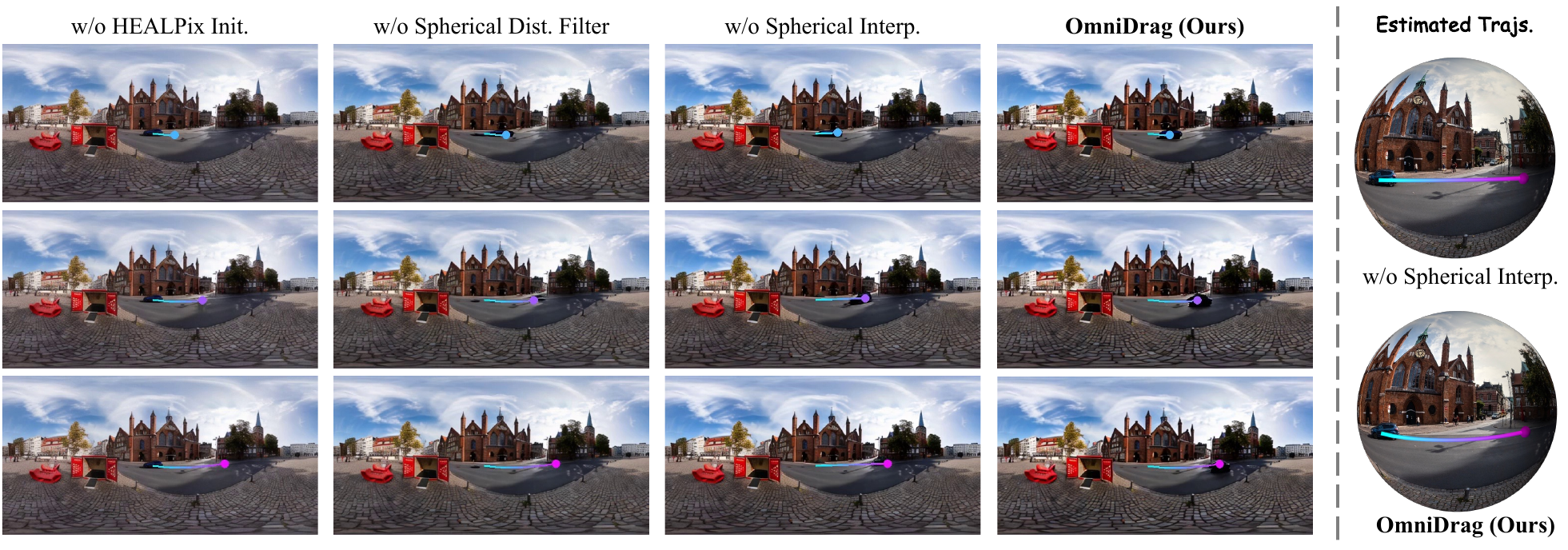}
    \vspace{-20pt}
    \caption{\textbf{Ablation study} on proposed spherical motion estimator (SME). The ``w/o HEALPix init." variant fails to control the car, the ``w/o spherical dist. filter" variant generates unstable result, and the ``w/o spherical interp." variant leads to unintended path. In contrast, our OmniDrag leverages SME to obtain precise and reasonable trajectories during training and inference, achieving pleasant results.}
    \label{fig:ablation_SME}
    \vspace{-12pt}
\end{figure*}

\textbf{Implementation Details.} We choose stable video diffusion (SVD) model \cite{blattmann2023stable} as our base model. We use CoTracker \cite{karaev2023cotracker} as the tracking function $\mathcal{F}_t$ and train the OmniDrag on Move360 and WEB360 \cite{wang2024360dvd} dataset. In the traning stage, we follow ReVideo \cite{mou2024revideo} to sample the number of trajectories $N_{samp}$ randomly between $1$ and $10$, and optimize OmniDrag with Adam optimizer \cite{loshchilov2017decoupled} for 40K iterations on $8$ A100 GPUs, with a batch size of $4$ for each GPU. The resolution is downsampled to $640\times 320$, the learning rate is set to $1 \times 10 ^{-5}$, and it takes about 2 days for training. Besides, we adopt a latent rotation mechanism \cite{wu2023panodiffusion, wang2024360dvd} to enhance the warp-around consistency of ODVs.

\noindent\textbf{Evaluation metrics.}  We follow MotionControl \cite{wang2024motionctrl} to evaluate from the following two aspects: (1) The quality of results is assessed using the Fréchet Inception Distance (FID) \cite{seitzer2020pytorch} and Fréchet Video Distance (FVD) \cite{unterthiner2018towards}, which measure the visual quality and temporal conherence, respectively. (2) The motion control performance (ObjMC) is evaluated by the spherical distance between the trajectories of the generated videos and the user input trajectories. In addition, we conduct a human evaluation, in which thirty volunteers are asked to vote the best method for each sample from two aspects: overall quality and motion matching.

\subsection{Comparison with State-of-the-Art Methods}
We compare our OmniDrag with state-of-the-art video generation methods incorporating motion control, specifically DragNUWA \cite{yin2023dragnuwa}, MotionCtrl \cite{wang2024motionctrl}, and DragAnything \cite{wu2025draganything}. We conduct experiments under both scene-level and object-level control conditions. Because MotionCtrl does not support trajectory control in image-to-video generation, we compare it only under scene-level control, using the corresponding camera pose as conditional input. We select ODIs from ODISR \cite{deng2021lau} and SUN360 \cite{xiao2012recognizing} datasets as reference images, and create twelve pairs of input as the test set. 

The visual comparisons of some cases are shown in Fig.~\ref{fig:comparison}, including scene-level control (top) and object-level control (bottom). Due to the lack of prior knowledge of spherical motion patterns, DragNUWA fails in both cases, producing only slight movements. MotionCtrl generates camera pose transformations in a 2D image manner, while DragAnything produces violent motions that distort image content. In contrast, our OmniDrag performs well in both scene-level and object-level control, conforming to the distortion and motion pattern of ODVs and exhibiting good warp-around continuity. More visual results are provided in the supplementary materials. Quantitative comparisons are presented in Tab.~\ref{tab:comparison}. Note that we also compare metrics on horizontal 8 viewports, which represent users' final content view. It can be seen that our OmniDrag achieves the best FVD on ERP format and the best FID and FVD on the horizontal eight viewports, demonstrating the good quality of our generated results. Notably, DragNUWA typically generates minimal motion, resulting in a lower FID on ERP but a poor ObjMC score, whereas our OmniDrag achieves superior motion consistency. Furthermore, in human evaluations, OmniDrag exhibits clear advantages over other methods, demonstrating its superior performance in video quality and instruction comprehension.

\begin{table}[!t]
\caption{\textbf{Ablation study} on five variants of OmniDrag.}
\vspace{-10pt}
\label{tab:ablation}
\centering
\small
\resizebox{1.0\linewidth}{!}{
\begin{tabular}{l|ccc|cc}
\shline
\rowcolor[HTML]{EFEFEF} &
  \multicolumn{3}{c|}{\small\cellcolor[HTML]{EFEFEF}ERP Image} &
  \multicolumn{2}{c}{\footnotesize\cellcolor[HTML]{EFEFEF}Horizontal 8 viewports}  \\ \hhline{>{\arrayrulecolor[HTML]{EFEFEF}}->{\arrayrulecolor{black}}|-----} 
\rowcolor[HTML]{EFEFEF} 
\multicolumn{1}{c|}{\multirow{-2}{*}{\cellcolor[HTML]{EFEFEF}Method}} & 
 \small FID$\downarrow$ &
 \small FVD$\downarrow$ &
 \small ObjMC$\downarrow$ &
 \small FID$\downarrow$ &
 \small FVD$\downarrow$ 
  \\ \hline \hline

{w/o. Ft Temporal Attn.} & 182.72 & 982.41 & 0.080 & 97.12 & 332.97\\ 

{w/o. Move360 Dataset}  & \best{167.56} & 941.58 & 0.327 & 95.82 & \best{317.73}\\ 
{w/o. HEALPix Init.} & \secondbest{170.69} & \secondbest{938.18} & 0.226 & \best{95.33} & 324.05 \\ 
{w/o. Shperical Filter.} & 174.40 & 970.06 & 0.113 & 96.31 & 336.81 \\ 
{w/o. Shperical Interp.} & 174.13 & 965.47 & \secondbest{0.053} & 96.94 & 342.18 \\

\textbf{OmniDrag (Ours)} & 171.41 & \best{933.73} & \best{0.044} 
& \secondbest{95.62} & \secondbest{322.22} \\ \shline
\end{tabular}}
\vspace{-10pt}
\end{table}

\subsection{Ablation Study}
To validate the effectiveness of the proposed components in OmniDrag, including the joint fine-tuning strategy, the SME, and the Move360 dataset, we conduct ablation studies, as shown in Figs.~\ref{fig:ablation_temporal} and~\ref{fig:ablation_SME}, and Tab.~\ref{tab:ablation}.

\textbf{Effect of jointly tuning temporal attention.} To demonstrate the importance of jointly tuning the temporal attention layers, we create a variant where we freeze the entire main UNet denoising branch.  The results in both ERP format and viewport are shown in Fig.~\ref{fig:ablation_temporal}.  It can be observed that ``w/o. fine-tuning" variant generates videos with only trivial 2D zoom-in effects, lacking omnidirectional properties,  which leads to distorted viewport quality. This variant also results in higher FID and FVD scores, as shown in Tab.~\ref{tab:ablation}.

\textbf{Effect of training on Move360 dataset.} We evaluate another variant by training our OmniDrag only on the WEB360 \cite{wang2024360dvd} dataset. Although this variant achieves better FID results, it exhibits poor motion control performance, as indicated in Tab.~\ref{tab:ablation}. The results in Fig.~\ref{fig:ablation_temporal} further illustrate that training without datasets containing high-quality motion cannot provide scene-level controllability due to insufficient motion diversity. In contrast, training with our Move360 dataset enables accurate and stable scene-level control, significantly enhancing the model's capabilities.

\textbf{Effect of SME.} To demonstrate the effectiveness of the proposed SME, we replace the HEALPix initialization, spherical distance calculation and spherical interpolation with 2D grid initialization, Euclidean distance and linear interpolation, respectively. The results are presented in Fig.~\ref{fig:ablation_SME} and Tab.~\ref{tab:ablation}. It is evident that removing these components significantly degrades the performance of motion control. Specifically, without HEALPix initialization and spherical distance filtering, the variant fails to control the object or generates unstable results, likely due to the lack of sufficient and accurate control signals during training. During inference, given the user's input of handle and target points, SME estimates a reasonable spherical trajectory, whereas the linear interpolation variant generates a straight line on the sphere, resulting in ambiguous results, \textit{e.g.}, the car runs off the road in this case, deviating from the intended path.

\section{Conclusion}
In this paper, we proposed \textbf{OmniDrag}, a novel diffusion-based approach for enabling motion control in omnidirectional image-to-video generation. We introduced an Omni Controller, which receives spherical trajectories as input, allowing for easy drag-style control. To effectively learn complex spherical motion patterns, we proposed jointly fine-tuning the controller and temporal layers in the diffusion denoising UNet. Additionally, we designed a spherical motion estimator to capture accurate control signals during training and provide user-friendly interaction during inference. Furthermore, we collected Move360, a new high-quality ODV dataset featuring significant motion content, which enhances OmniDrag's scene-level controllability. Experiments manifested that OmniDrag achieves state-of-the-art performance in both scene- and object-level motion control.

\noindent\textbf{Limitations.} Although OmniDrag achieves promising results, its generation quality is constrained by the base SVD model in certain scenarios. Moreover, decoupling camera- and object-level motion is an open problem for future work.

\small
\bibliographystyle{ieeenat_fullname}
\bibliography{ref}
\clearpage
\maketitlesupplementary
\setcounter{section}{0}
\renewcommand\thesection{\Alph{section}}

Our main paper has outlined the core techniques of our proposed \textbf{OmniDrag} method to enable motion control for omnidirectional image-to-video generation. It has also demonstrated the efficacy of our methodological contributions through experiments. This appendix offers further details on our Omni Controller in Sec.~\ref{sec:supp_A}, more details of our Move360 dataset in Sec.~\ref{sec:supp_B}, along with additional experimental results and analyses in Sec.~\ref{sec:supp_C}, which are not included in the main paper due to space constraints. We also provide   a \href{https://lwq20020127.github.io/OmniDrag/}{ project page}
 to show more results, together with more sample videos from our Move360 dataset.

\begin{figure}[!t]
    \centering
    \includegraphics[width=1.0\linewidth]{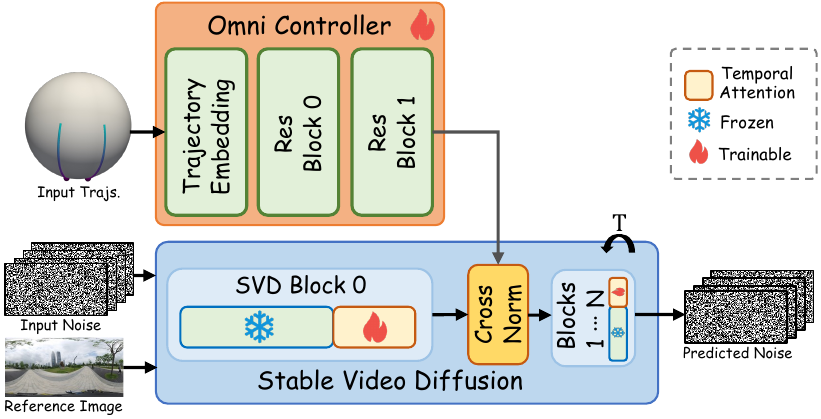}
    \vspace{-16pt}
    \caption{Illustration of our Omni Controller. Our Omni Controller only consists of a trajectory embedding module and two ResBlocks. The output control signal is integrated into the main SVD branch at its first block (SVD Block 0) by adding them to the denoising feature after applying the cross normalization.}
    \label{fig:supp_omnicontroller}
    \vspace{-12pt}
\end{figure}

\section{More Details of Omni Controller}
\label{sec:supp_A}
An illustration of our Omni Controller is shown in Fig.~\ref{fig:supp_omnicontroller}. We further elaborate on the details of Omni Controller in the following.

\noindent\textbf{Trajectory Embedding.} Recall that each sampled trajectory $\mathbf{T}_j\in \mathcal{T}'$ is defined as a sequence of spatial positions $\mathbf{T}_j=\{(x^i_j, y^i_j)|i\in \{0, 1, \dots, L-1\}\}$, where $(x_j^i, y_j^i)$ represents the position of the $j$-th trajectory at frame $i$. We follow MotionCtrl~\cite{wang2024motionctrl} to explicitly expose the moving speed of the object, as:
\begin{equation}
\begin{aligned}
    \{\left(0,0\right), \left(u_{(x_1,y_1)}, v_{(x_1,y_1)}\right),\dots,\left(u_{(x_{L-1}, y_{L-1})}, v_{(x_{L-1}, y_{L-1})}\right)  \},
\end{aligned}
\end{equation}
where $u_{(x_i, y_i)}=x_i-x_{i-1}$, $v_{(x_i, y_i)}=y_i-y_{i-1}$, and $i\in \{0, 1, \dots, L-1\}$. The first frame and the other spatial positions in the subsequent frames that the trajectories do not pass are denoted as $(0,0)$. As a result, $\mathbf{T}'\in \mathbb{R}^{L\times H \times W \times 2}$, where $H$ and $W$ are the height and width of the input ODV, respectively. A Gaussian filter is then applied to smooth the sampled trajectories, and some convolution blocks are adopted to upsample the channel dimension to $320$.  Finally, the condition $\mathbf{c}\in \mathbb{R}^{L\times H \times W \times 320}$.

\noindent\textbf{Control Injection.}
We follow ControlNeXt~\cite{peng2024controlnext} to use a lightweight architecture only composed of two ResBlocks~\cite{he2016deep}. This pruning maintains the model's consistency while significantly reduces latency overhead and parameters. For injection of the control signals, we use cross normalization technique (Eq.~\ref{eq:cross_norm} in the main paper), which aligns the distribution of the denoising and control features, serving as a bridge to connect the diffusion and control branches. Finally the normed control is integrated into the main branch at the first block (SVD Block 0 in Fig.~\ref{fig:supp_omnicontroller}) by addition.

\section{More Details of the Move360 Dataset}
\label{sec:supp_B}
We have provided the camera parameters and equipment used for data acquisition in the main paper. Here, we detail our data selection strategy and process based on three key criteria:
(1) \textbf{Scene categories:} The dataset should encompass a wide range of scenes, including schools, parks, markets, landscapes, and more. Besides, the video should not have large-scale obstruction from people or buildings.
(2) \textbf{Lighting conditions:} The dataset should cover various lighting conditions, such as indoor and outdoor settings, daytime (sunny and cloudy), and nighttime.
(3) \textbf{Motion magnitude:} To enhance motion controllability, the videos in the dataset should exhibit relatively large motion magnitudes, avoiding examples where the background scene and object are all static.

\begin{figure*}[!t]
    \centering
    \includegraphics[width=1.0\linewidth]{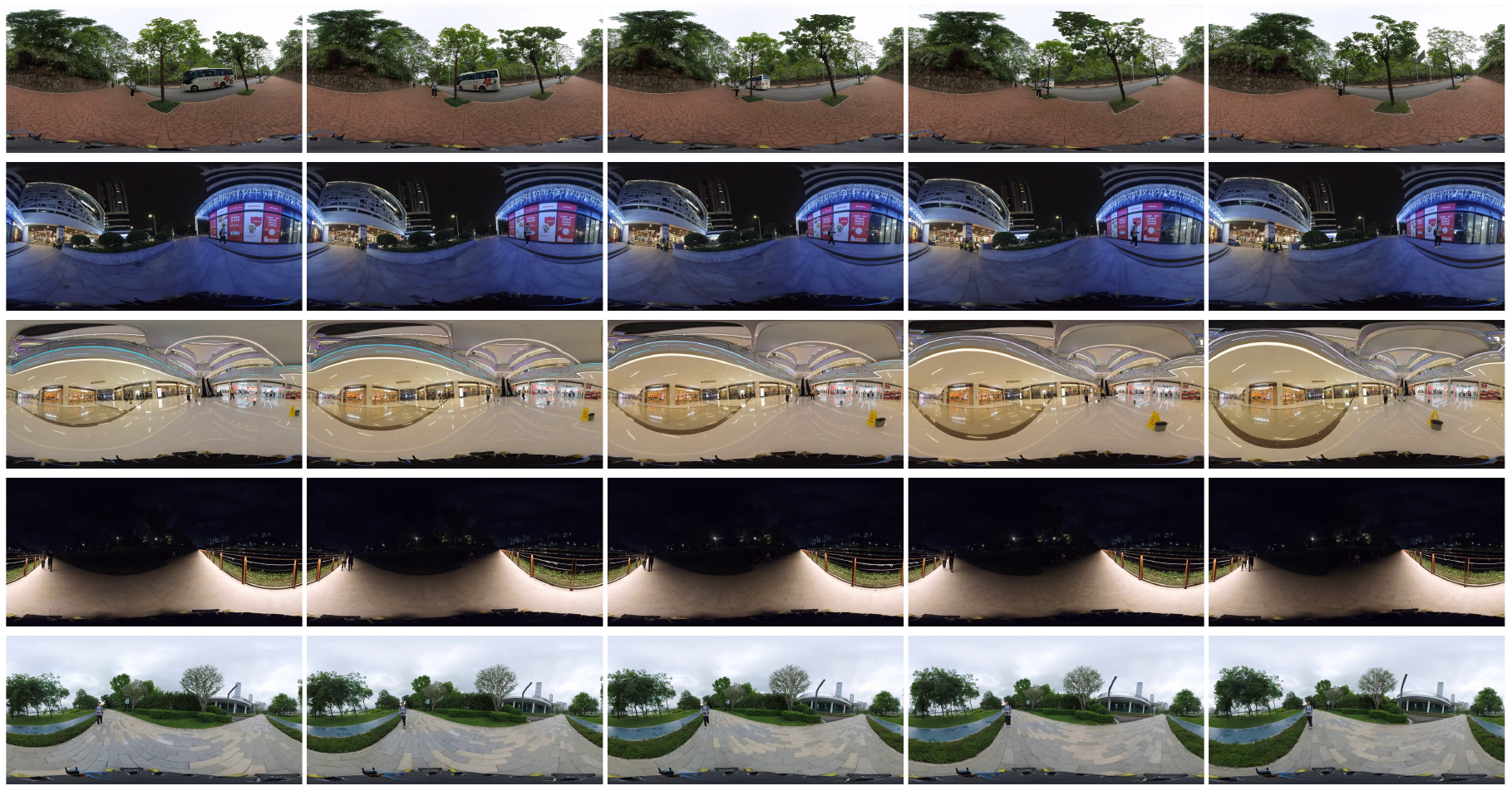}
    % \vspace{-14pt}
    \caption{Some sample videos in our Move360 dataset.}
    \label{fig:supp_dataset}
    % \vspace{-12pt}
\end{figure*}

Specifically, the original video footage has a duration of approximately 20 hours and a size of 6~TB. We first employed optical flow to filter out video segments that are nearly stationary. We then evenly split the remaining video and manually reviewed each clip, considering scene category, lighting conditions, and image quality, and excluded clips with device debugging or occlusions, resulting in 2,100 clips, each consisting of 100 frames. Finally, we utilized CoTracker~\cite{karaev2023cotracker} to calculate the average spherical motion distance of each video and filtered out the lowest 25\% based on this metric. After double-checking the filtered videos, we finalized a dataset of 1,580 clips. Fig.~\ref{fig:supp_dataset} presents samples from the Move360 Dataset. We also provide a \href{https://lwq20020127.github.io/OmniDrag/}{ project page} to showcase more video samples. We hope that the Move360 Dataset will help fill the gap in the field due to the lack of large-scale motion video datasets. We believe that Move360 will serve as a valuable resource for advancing research in omnidirectional video technologies, fostering innovation and collaboration within the community.

\section{More Experiment Results}
\label{sec:supp_C}
\begin{figure*}[!t]
    \centering
    \includegraphics[width=1.0\linewidth]{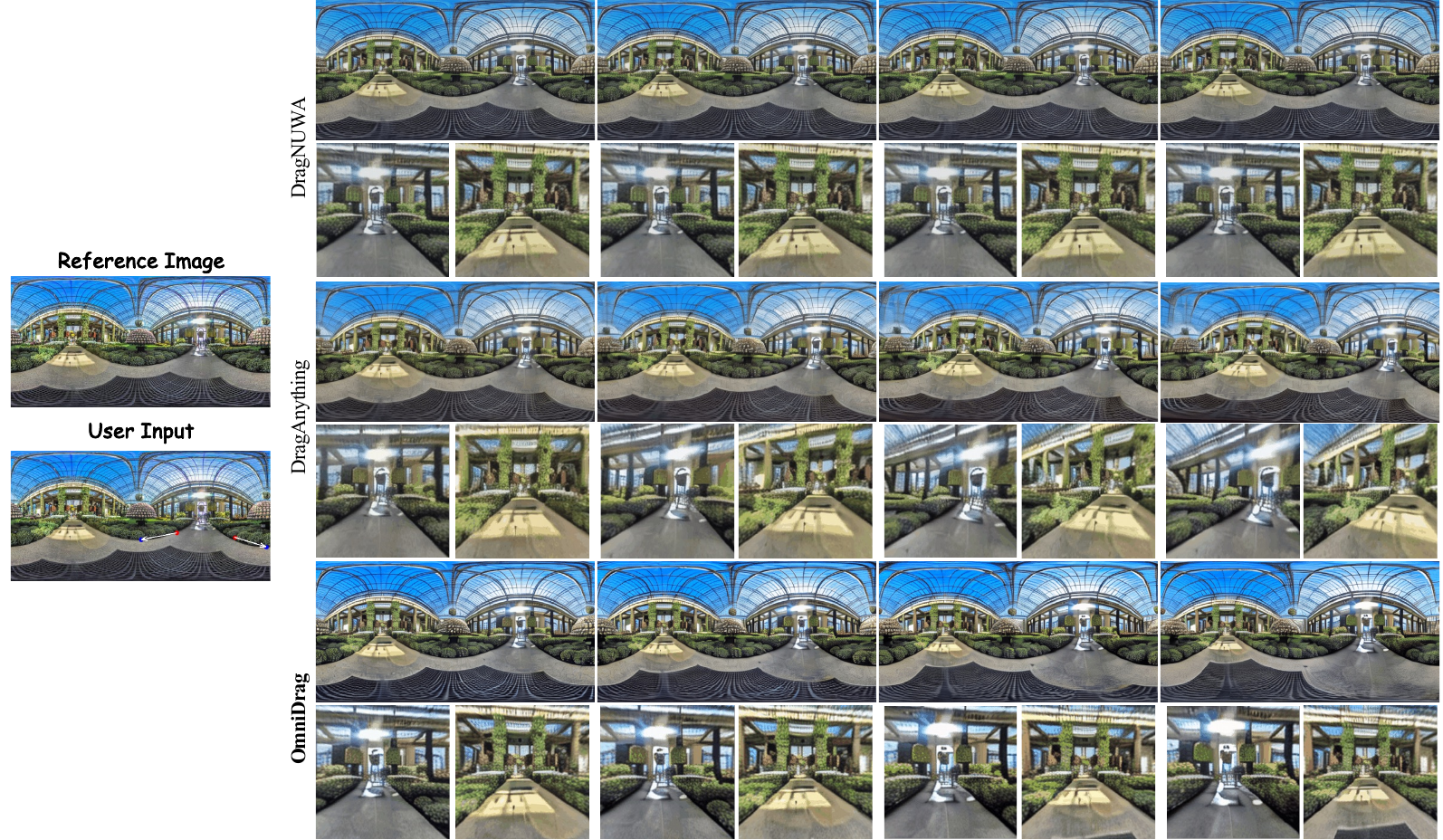}
    % \vspace{-14pt}
    \caption{\textbf{Visual comparisons} of between DragNUWA \cite{yin2023dragnuwa}, DragAnything \cite{wu2025draganything}, and our OmniDrag. For each ERP image, we show two corresponding viewports at specific perspectives.}
    \label{fig:supp_exp_1}
    % \vspace{-12pt}
\end{figure*}

\begin{figure*}[!t]
    \centering
    \includegraphics[width=1.0\linewidth]{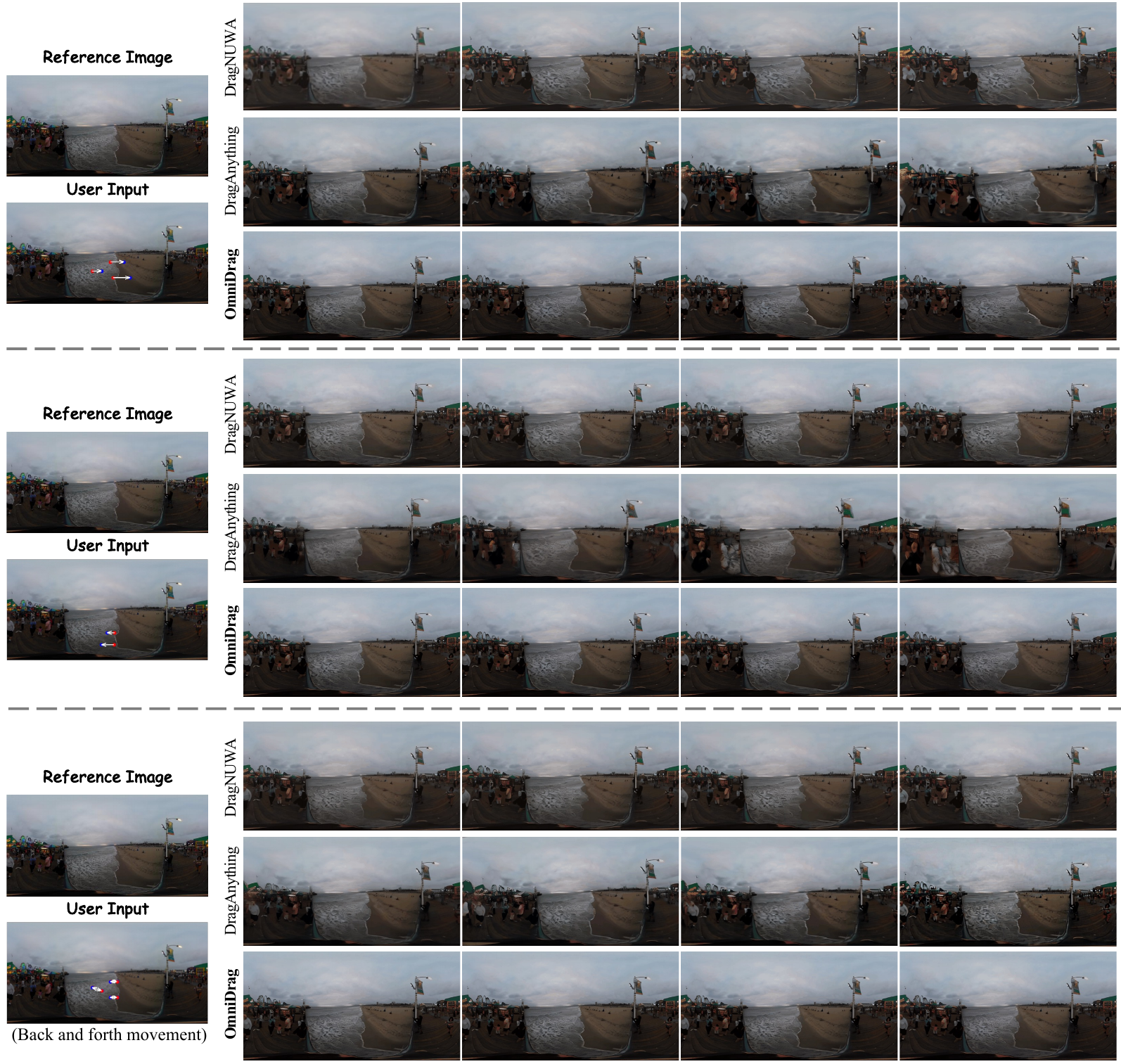}
    % \vspace{-14pt}
    \caption{\textbf{Visual comparisons} of between DragNUWA \cite{yin2023dragnuwa}, DragAnything \cite{wu2025draganything}, and our OmniDrag on the same reference image under different drag controls.}
    \label{fig:supp_exp_2}
    % \vspace{-12pt}
\end{figure*}

Additional visual comparisons are provided in Figs.~\ref{fig:supp_exp_1} and~\ref{fig:supp_exp_2}. In Fig.~\ref{fig:supp_exp_1}, the control condition is the inversion of the case presented in Fig.~\ref{fig:comparison} of the main paper; two rendered viewports at specific perspectives are also included. It can be observed that OmniDrag achieves stable scene-level control, whereas DragNUWA generates only artifacts. Furthermore, DragAnything distorts the ERP distribution, resulting in deformed user viewports. Figure~\ref{fig:supp_exp_2} illustrates different drag inputs applied to a single reference image. DragAnything and DragNUWA also affect background regions not intended for control, whereas OmniDrag achieves precise object-level control. Additional results, including those in ERP format and various viewports, are provided on our  \href{https://lwq20020127.github.io/OmniDrag/}{ project page}.

\clearpage

% \bibliographystyle{ieeenat_fullname}
% \bibliography{ref}
\end{document}